\documentclass[runningheads]{llncs}

\usepackage{graphicx}

\usepackage{amssymb}
\usepackage{amsmath}
\usepackage{url}

\usepackage{array}
\usepackage{changepage}

\usepackage{comment}

\usepackage{bbold}

\DeclareMathOperator*{\argmax}{argmax}

\begin{document}
\title{Confident AI} 
\titlerunning{Confident AI}

\author{Jim Davis}
\authorrunning{J. Davis}

\institute{Dept. Computer Science and Engineering\\
Ohio State University\\
Columbus OH 43210 USA\\
\email{davis.1719@osu.edu}}

%
%
\maketitle              
\begin{abstract} 
In this paper, we propose ``Confident AI'' as a means to designing Artificial Intelligence (AI) and Machine Learning (ML) systems with both algorithm and user confidence in model predictions and reported results. The 4 basic tenets of Confident AI are Repeatability, Believability, Sufficiency, and Adaptability. Each of the tenets is used to explore fundamental  issues in current AI/ML systems and together provide an overall approach to Confident AI. 
\end{abstract}
%
%
%

\section{Introduction}

Artificial Intelligence (AI) has many definitions, types, and themes that have appeared to describe and promote various aspects. Two broad {\em definitions} of AI include ``the scientific understanding of the mechanisms underlying thought and intelligent behavior and their embodiment in machines'' \cite{AAAI} and ``AI strives to build intelligent entities as well as understand them” \cite{RussellNorvig}. Three primary {\em types} of AI include Narrow, General, and Artificial Superintelligence. Narrow (or Weak) AI includes approaches that are task oriented and limited in scope (where we mostly are today).
General (or Strong) AI seeks to achieve a human-level ability to solve broad problems (how we would like it). Lastly, Artificial Superintelligence has the goal of surpassing human intelligence (do we even want this?). Furthermore, multiple AI {\em themes} have recently appeared, such as Explainable AI, Trustworthy AI, and Ethical/Beneficial/Responsible AI. These themes seek to develop AI with transparency and respect of privacy (Explainable AI, Trustworthy AI) along with adherence to ethical guidelines and values to build systems beneficial to humans (Ethical/Beneficial/Responsible AI). Machine Learning (ML), which includes those techniques for analyzing and drawing inferences based on patterns automatically learned from data, is also considered part of the overall scope of AI.

In this paper, we introduce a new complimentary theme which we call ``Confident AI". To provide an initial motivation, we refer to a recent article \cite{Tucker2021} that describes an evaluation of a specific US Air Force target classification model:
 \begin{quotation}
…the low accuracy rate of the algorithm wasn’t the most worrying part of the exercise. While the algorithm was only right 25 percent of the time, he said, ``It was confident that it was right 90 percent of the time, so it was confidently wrong.''
 \end{quotation}
 
\noindent Of primary concern here is that the deployed algorithm reported very high confidence in its decisions (90\% success rate) even though it was wrong most of the time (correct only 25\% of the time). Clearly such overconfidence is undesirable. This example provides a clear case when the algorithm's confidence scores did not reflect actual performance, and thus resulted in a total lack of confidence in the system by the user.

To combat such issues, we present Confident AI as a means to develop  AI systems that have both {\em internal} (algorithm) and {\em external} (user) confidence on the output decisions and predictions.  The internal confidence provides reliability in the results that can be used in further automated downstream tasks, while the resulting external confidence helps the end-user build trust in the system. In the next section, we will explore various confidence related issues that may occur and approaches on how to avoid them. 


\section{Tenets of Confident AI}

We propose 4 basic tenets of Confident AI: Repeatability, Believability, Sufficiency, and Adaptability. These principles can be used to enable various AI/ML systems to operate with confidence. We employ multiple classification examples throughout to illustrate the concepts, though the ideas are transferable to other decision-making paradigms.  The proposed tenets can be applied separately to provide useful benefits, and together they unify to yield a general approach to Confident AI.
 

\subsection{Repeatability}

The typical experimental pipeline when proposing a new model/approach is to train the model, test the model, report the results, and compare the results against other methods. Often times very slight differences in results between methods are reported (e.g., $\pm$1\% difference in accuracy) and are used to claim success of one approach over another. However,  results using only a single training run of a model are {\em not} sufficient to capture the underlying variation in performance that is possible. 

For a given/fixed model architecture and training regimen, performance variability arises from simply applying a different random seed initialization (affecting initial model parameter values and ordering of training data) \cite{Picard2021,DavisFrank2021} or from using different computing hardware (different GPUs). For example, multiple training runs of a hypothetical classifier producing an average test accuracy of $\mu$=85\% with a standard deviation of $\sigma$=1\% could therefore have an observable accuracy range from 82-88\% (for $\pm3\sigma$) based solely on the effects due to randomized training initializations. Thus results presented or compared from just a single (perhaps even best) training run are not strictly repeatable.  Instead, a statistical measurement across multiple training runs is required to properly evaluate and compare different methods.

Though the field is beginning to see more papers reporting the mean, and sometimes standard deviation, of scores from a small number (e.g., 3) of random initialization training runs, this unfortunately still leaves it to the reader of the paper to subjectively assess any meaningful difference in the compared methods. In our work of \cite{DavisFrank2021}, we collected test results from 15 randomly trained networks of each model for each experiment. Then a one-sided paired T-Test \cite{Devore2011} (using a p-value of .05) was employed to statistically prove if an improvement with the proposed model exists over other methods. We assert that such a statistical approach, common to other scientific fields, is required in AI/ML to properly evaluate and compare results. 

Of particular note is that there is a distinction between “statistical” significance and “practical” significance. It may be the case that results of one method are statistically better than another, however the amount of difference may be slight (e.g., $<$1\% in accuracy) and have no real practical benefit in deployment. In such cases, other aspects such as computational cost will likely be the dominate factor in choosing which model is ``better''. 

So far, a procedure for repeatability using statistical evaluations has been proposed to summarize and compare results, but how to deploy the chosen method still remains. Various options include selecting the single best performing model from the multiple training runs, keeping the entire collection of models for use as an ensemble \cite{Bishop2006}, or distilling the ensemble back to a single model \cite{Hinton2015}. The particular deployment scenario will likely dictate which option is most appropriate. 

Multi-run training and statistical evaluation will certainly take additional time and resources. However, we argue that such a cost is important for Confident AI to account for the underlying variabilities in training and the overall repeatability of model evaluation and comparison.


\subsection{Believability}

To make a prediction/decision based on the input data, the standard maximum {\it a posteriori} (MAP) approach is to select the hypothesis having the largest posterior probability among competing hypotheses. For a Neural Network, this corresponds to (argmax) selecting the class having the highest output softmax or logit value. But what if that top prediction is uncertain? How can we ascertain its actual confidence? 

One needs to be able to assign a reliable confidence score as a measure-of-belief to any output prediction.
Such scores are deemed to be ``perfectly calibrated'' if they correspond to the “probability of being correct” \cite{Guo2017}. For example, a calibrated decision-making system giving a confidence score of~.9 for a classification on a particular input would accurately reflect that the decision has a 90\% chance of being correct. An \underline{un}calibrated score of~.9 could therefore be over-confident (in reality is $<$90\% of being correct) or under-confident (actually is $>$90\%). Given a {\em collection} of prediction confidence scores, the decision approach could be categorized overall as being well-calibrated, over-confident,  under-confident, or simultaneously both over- and under-confident (across different classes or ranges of scores). 
In modern Neural Networks, the output softmax values themselves are typically not expected to be well calibrated. However there exist multiple methods to measure the calibration quality and to re-calibrate the system \cite{Guo2017}. 

\begin{figure}[!t]
\centering
\setlength{\tabcolsep}{0.0pt}
\begin{tabular}{ccc}
\includegraphics[height=1.5in]{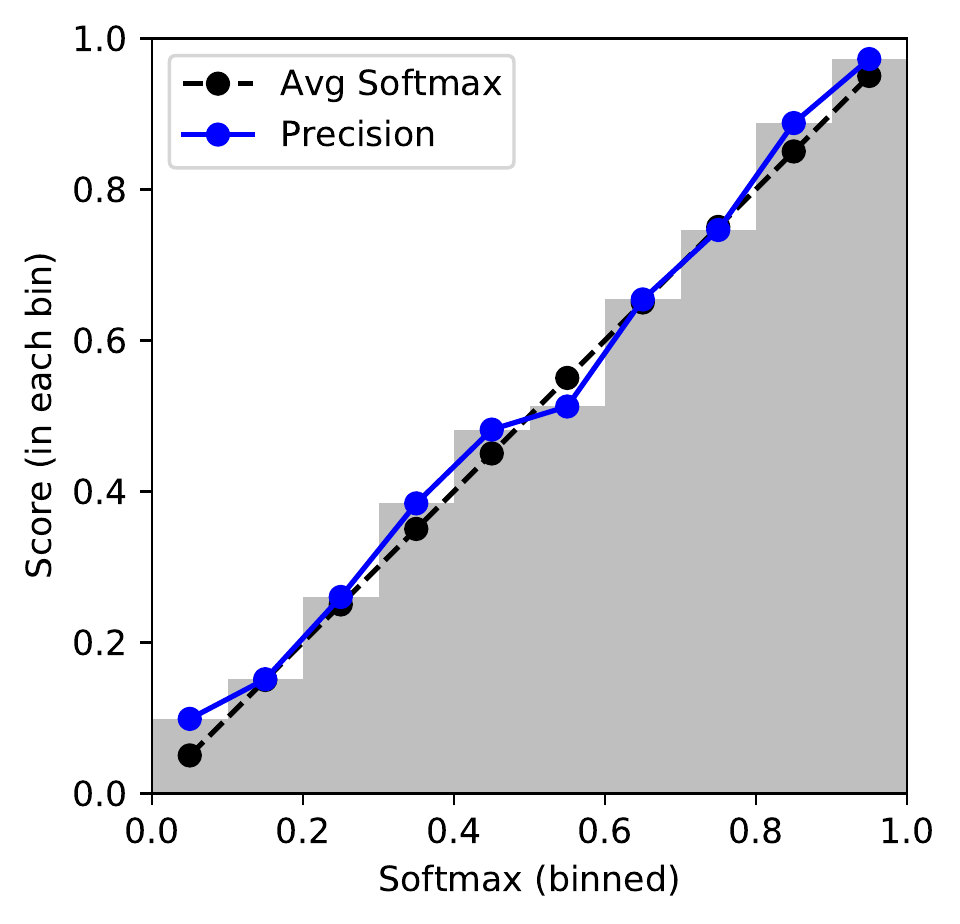} & 
\includegraphics[height=1.5in]{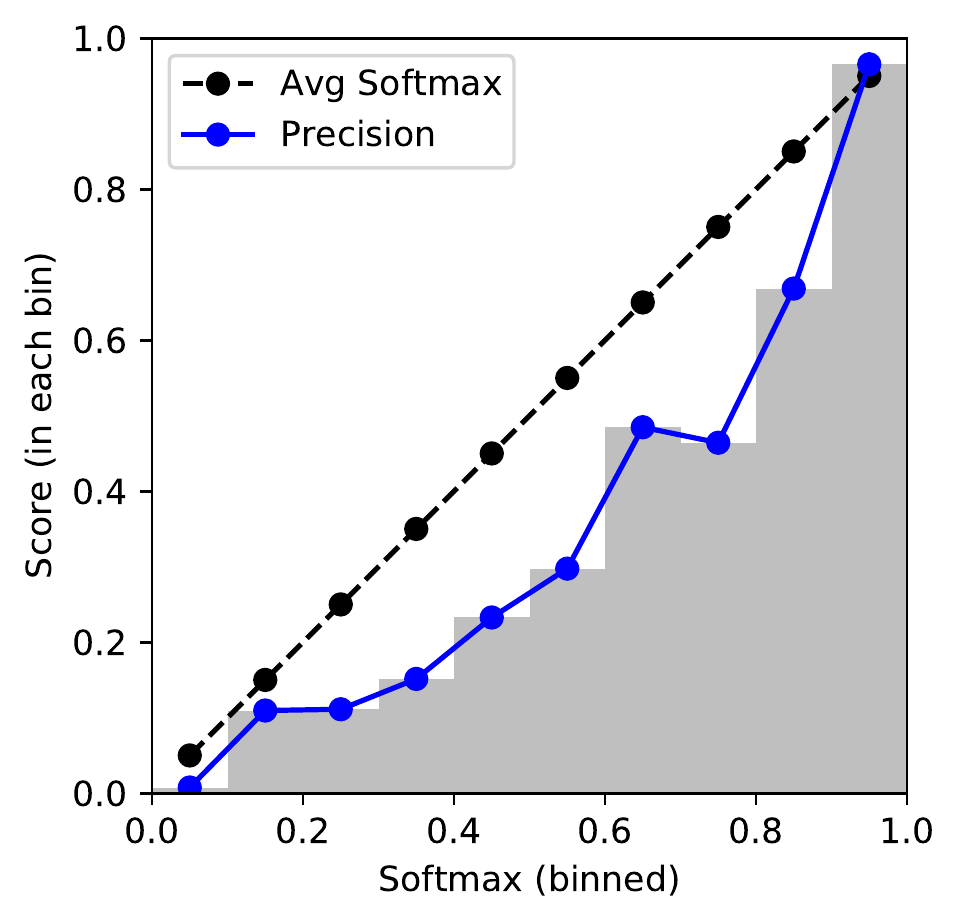} & 
\includegraphics[height=1.5in]{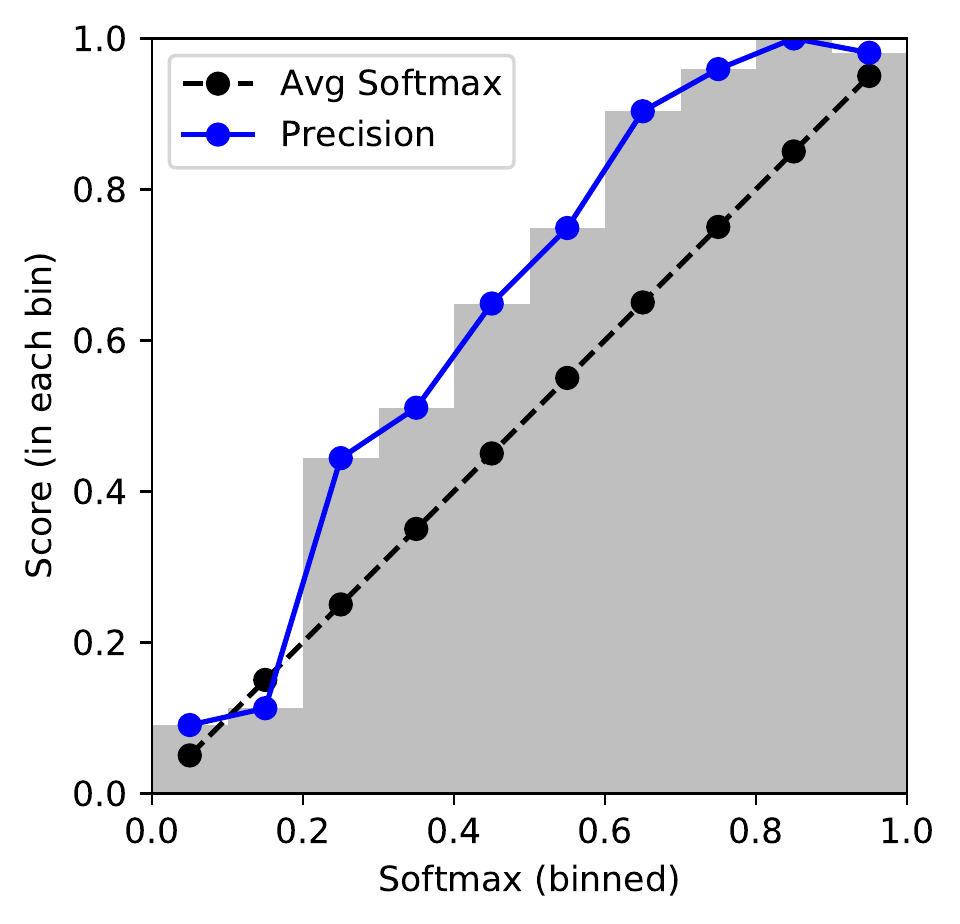} \\
Well-calibrated & Over-confident & Under-confident
\end{tabular}\caption{Reliability diagrams.}
\label{fig:reliability-diagrams}
\end{figure}

Reliability diagrams offer a useful {\em qualitative} assessment of the calibration by binning/quantizing the argmax-selected softmax values for a set of validation examples and computing the precision of the classifications in each bin (proportion of correct classifications in the bin). A plot of the per-bin  average softmax and precision can be used to provide visual assessment of the calibration quality across the range of argmax-selected softmax values. Example reliability diagrams showing well-calibrated, over-confident, and under-confident classifiers are shown in Fig.~\ref{fig:reliability-diagrams}. An ideally calibrated classifier would have precision bin heights closely following the corresponding average softmax in each bin (dashed line). Measured precision values in bins that are {\em below} the dashed line signify over-confidence, as the softmax values are higher than the corresponding precision. Conversely, bins with precision values {\em above} the dashed line demonstrate under-confidence.

A {\em quantitative} measure of the calibration based on the reliability diagram is Expected Calibration Error (ECE), 
\begin{equation}
\text{ECE} = \sum_{m=1}^M \frac{|B_m|}{N} \left| \text{avg-softmax}(B_m) - \text{precision}(B_m) \right|
\end{equation}

\noindent which is the weighted average (using the proportional number of examples in each bin) of the absolute difference between the average softmax and precision in each bin $B_m$. This measurement attempts to summarize the reliability diagram into a single score. A small ECE value is desired, otherwise the prediction confidence (softmax) cannot be trusted. 

Ideally, measuring calibration should extend to all of the output softmax values (for each possible class) for a given input, rather than just for the top, highest softmax value of each example. A form of {\em total} calibration error could be formulated similar to ECE, where the bins are populated with {\em all} of the output softmax values, not for just the argmax-selected class. Then, instead of using the precision per bin, the proportion of {\em ground-truth} classifications existing in each bin could be employed.

Beyond visualizing and measuring the calibration quality, post-processing methods are available to improve the calibration of a given classifier without changing the final classification results \cite{Guo2017}. Most popular is the technique of temperature scaling, where the pre-softmax logits are divided by a constant value (temperature) $T$ which has a value chosen to either flatten  ($T>1$) or sharpen ($T<1$) the resulting softmax distribution. Adjusting the softmax distribution in such a way can considerably improve the calibration while maintaining the relative ordering of the softmax values in the distribution (thus the top class remains unchanged). A proper value of $T$ for optimizing calibration on validation data can be learned or found via grid-search \cite{Guo2017}. 

Calibration techniques can also operate in a global or per-class manner. In per-class calibration, all of the argmax-selected predictions of a given class are calibrated independently of all other class predictions. During inference, a specific per-class calibration setting ($T_{class}$) is recalled and applied to the logits based on the corresponding classification of the  prediction. The main advantage of per-class over global calibration is that per-class is better able to handle separate under- or over-confidence for different classes \cite{Zhao2020}.

The result of employing calibration is an increased believability in the decision architecture's output confidence. Further downstream tasks and user/analyst acceptance (trust) are therefore improved when receiving well-calibrated posterior confidence scores.  We further desire not to give highly confident predictions at test/inference time to ``unwanted'' examples, such as out-of-domain/out-of-class or adversarial examples, and thus extensions with Open Set Recognition \cite{Geng2021,Mahdavi2021} and adversarial detection techniques \cite{Aldahdooh2022} are also of importance to include in real-world scenarios.


\subsection{Sufficiency}

Given a well-calibrated decision mechanism, it may be necessary in specific tasks or mission-critical applications for predictions to meet some sufficiently high level of confidence before any resulting action can be taken.  With such a condition, what is the appropriate means to handle those predictions that fall {\em below} the required confidence level? We propose two approaches to address below-confidence predictions:  rejection and generalization. In a rejection scheme, a method is used to detect and reject unconfident predictions. The rejected examples are then either discarded or re-assessed by a different means, perhaps with a more complex model using additional data/sensors or sent to a human analyst to resolve. Rather than rejecting a problematic example, an alternative strategy is to repeatedly generalize or ``soften'' the prediction's semantics to increase the confidence until reaching the required confidence threshold.  Such an approach is possible when the decision space can be hierarchically composed.


\subsubsection{Rejection.}

Standard Neural Network classifiers typically ``force'' classification by choosing the argmax class (the class having the largest softmax/logit) as the final output decision. However, we argue that for Confident AI, a model should know/learn when to say ``I don’t know'' when given a confusing or corrupted input. One simple approach is to just threshold the softmax value of the argmax class. For a perfectly calibrated model, one could choose a confidence threshold of 50\% (undecided) and remove any argmax classification having a softmax value $\leq$50\%. However, even in well-calibrated models, a 50\% threshold may not be the optimal choice, as optimality can be specified in a variety of ways. A better approach to determine the threshold can be formulated using the techniques of ``reject option classification'' \cite{Franc2021}. Reject option classifiers are a form of constraint-based decision making techniques where an alternative {\sc reject} class is added to the list of possible output classes. Underlying the accept/reject threshold mechanism is a constrained optimization task typically based on the final ``coverage'' (proportion of examples accepted) and  ``risk'' (error rate of accepted examples) after employing the threshold on a validation set. 

In a coverage-constrained approach, a threshold function is sought that minimizes the risk of accepted examples while constraining the coverage of accepted examples to meet a given percentage of the data. For example, at least 80\% of the examples may be required to be accepted while trying to minimize the classification error of the accepted examples. This approach can be applicable when a limited number of human analysts (or other costly processing methods) are employed to examine rejected (problematic) examples and thus the coverage constraint ensures that the analysts are not overloaded. However, many accepted examples could still be unconfident and produce errors. 

Another approach is to swap the coverage and risk in the optimization to instead maximize the coverage of accepted examples while constraining their risk. This method is based on the need for a given error rate of the classifier on the accepted examples. Use of this approach relates to mission-critical decision making where only a certain percentage of mistakes on accepted examples can be tolerated. One issue with this approach is that it may not be possible with a given model and dataset to reach the specified  error rate (risk) without having to remove a large portion (or most) of the data.

Depending on any task applicable constraints for risk or coverage, these reject option approaches can be used to find a threshold to identify and filter out confusing regions in the decision space. Then any  rejected examples could be discarded or re-assessed by other means.

\subsubsection{Generalization.}

In a situation where it is not desirable to reject a decision due to its low confidence, one possibility is to instead generalize or soften the decision to a semantically broader category which inherently would have more confidence. This type of re-classification is possible when the decision space has a natural (semantic) or user-specified hierarchical structure (e.g., a biological taxonomy used in a medical diagnosis). There exists a rich computational history in hierarchical analysis and processing, including  top-down (e.g., \cite{Wu2020,Lee2018,Liang2018}) and bottom-up (e.g., \cite{Deng2012,Davis2019,Davis2021}) classification approaches for different scenarios.  A  bottom-up framework can be used to repeatedly generalize and re-classify a terminal-level prediction (from the classifier) {\em upward} through the hierarchy until a particular confidence threshold is met. Thus the deepest (most specific) label having the required confidence is chosen. The overall goal is to aggregate sufficient confidence from relevant terminal-level classes into a super-class that still retains relevant semantic information about the original prediction, but with higher confidence.

For example, given a hierarchy of various vehicles (including types of cars, motorcycles, buses, trucks, etc.), a base prediction of “{\sc Toyota sedan}” having a  confidence of 61\%, and  a confidence requirement of at least 90\%, the original prediction could hypothetically be generalized to {\sc car} along the upward path as
\begin{center}	
{\sc Toyota sedan} (61\%) $\rightarrow$ {\sc sedan} (84\%) $\rightarrow$ {\sc car} (98\%) $\checkmark$
\end{center}

\noindent where the super-class {\sc sedan} may include many specific sedans (e.g., {{\sc Toyota sedan}, {\sc Honda sedan}, {\sc Lexis sedan}, etc.) and {\sc car} may broadly contain different body styles (e.g., {\sc sedan}, {\sc coupe}, {\sc hatchback}, etc.). Rather than discarding the original prediction of {\sc Toyota sedan} due to insufficient confidence, the bottom-up generalization process still provides a meaningful, yet softened, confident prediction of {\sc car} while ruling out other possibilities from {\sc motorcycle}, {\sc bus}, {\sc truck}, etc. In some cases, the generalization process may need to continue all the way up to the root of the hierarchy to meet a particularly high confidence threshold or to deal with a truly problematic input (e.g., corrupted data) and thus return ``{\sc unknown}''. This ability to determine that an input is unclassifiable is important.

With a classifier having perfect ``total'' calibration, where {\em all} of the output softmax values  are calibrated (not just the argmax softmax), evaluating any super-class in the hierarchy would simply require summing of the individual softmax values of its terminal class descendants and comparing the sum to the confidence threshold. However, it can be difficult to totally calibrate a classifier. Instead, our generalization approach described in \cite{Davis2019} employs a Bayesian posterior probability for each class in the hierarchy using known priors and likelihoods estimated from normalized histograms of the softmax value for the ground-truth class in positive/negative examples. An estimated softmax value for a non-terminal class is computed from the sum of softmax values (probability mass) of its terminal descendants. At inference time, the argmax-selected class from the base classifier is repeatedly generalized {\em upward} until a posterior meets the confidence threshold. Thus any originally correct terminal-level (initial) prediction remains on the correct upward path and any incorrect base prediction has the potential to be corrected at a valid super-class (an ancestor of the ground-truth terminal class).  This approach was applied to the task of hierarchical classification of cultivated plant stresses for quick and effective treatment \cite{Frank2021}. In our latter approach of \cite{Davis2021}, a formulation is derived using ``generalized logits'' instead of aggregated softmax values and employs an efficient logistic regression posterior model using generalized logits that span relevant hypotheses. This method is applicable to any logit/softmax-based classifier and provides a monotonic, non-decreasing hierarchical inference guarantee. 

In a situation when a pre-defined class hierarchy is unavailable, a similar process using {\em subsets} of terminal classes can be beneficial \cite{Davis2021}. A subset generalization approach is conducted by repeatedly adding terminal classes in descending order of their prediction confidence (from a totally calibrated classifier) until their combined probability exceeds the confidence threshold. In relation to the previous example where {\sc Toyota sedan} was generalized to {\sc car} due to a 90\% confidence threshold, a subset of the most confident terminal-level classes could be grown until reaching the required confidence (e.g., \{{\sc Toyota sedan} (61\%), {\sc Honda sedan} (23\%), {\sc Toyota coupe} (9\%)\} = 93\%).  This form of subset prediction can still offer utility in further processing or analyst tasks, and there would exist a high probability that the actual terminal-level answer is contained within the subset.

Overall, the notion of sufficiency and the ability to say ``{\it I don't know}'' or ``{\it I'm not sure}'' for confusing or corrupted data is important for critical applications in which high confidence is required to act on any given prediction. A rejection-based approach can be used to initially identify and filter out unconfident regions of the decision space, while a generalization-based method could be applied to soften an initial prediction until reaching a desired confidence level. 


\subsection{Adaptability}

Once a trained prediction model is deployed, it must be resilient to changes in the data landscape over time. It may be common to observe a transition in the data features (Domain/Distributional Shift), a change in the occurrence of certain classes (Prior Shift), and/or a modification of the input-output predictive relationship (Concept Drift). Therefore, a {\em static} prediction mechanism can quickly degrade in performance within a dynamic environment. A model must be able to properly adapt over time to ensure that it remains relevant in new situations. In general, various Continuous Learning/Training frameworks \cite{Parisi2019} may be used to address the aforementioned issues by repeatedly updating the model over time using new data. One key component to such model updating is {\em monitoring}, where a verification process of the current model is used to signal whether an update of the model is needed. Such ``lifelong learning'' helps to adapt models as situations change, but special care must be taken as not to completely undo what has already been learned. 


Consider a case of Prior Shift for a mobile robot that can repeatedly move between indoor and outdoor environments. Given a large set of objects for the robot to visually identify (classify) in each environment, the class priors of the objects will necessarily be different depending on which environment the robot is currently located. Therefore the robot's object classification model will need to be properly adjusted whenever a change between indoor and outdoor occurs, as detected by its monitoring procedure. With posterior probability assessments given by the visual classification system, a way to adapt to the change in priors is specifically needed.

Neural Network (and other classification) models are typically trained on well-balanced datasets  when available (i.e., with equal class priors) to produce posterior probabilities on the decision space. Similarly, approaches such as K-Nearest Neighbors can be used to compute local density estimates of the training examples/features to provide a posterior probability estimate of each class. Consequently, the initial class priors in these approaches are inherently part of the training process and thus affect the final classification. Rather than perform a costly retraining process when Prior Shift appears, a means to decouple the original priors from the existing output posteriors would enable estimation of new context-dependent posteriors using the new priors.  

In \cite{Davis2020}, we showed that given Bayes' Rule of the posterior $p(c_i | x)$ trained using priors $p_{train}(c_i)>0$ with $N$ classes,
\begin{equation}
p(c_i | x) = \frac{p(x | c_i)  p_{train}(c_i)}{\sum_{j=1}^N p(x | c_j) p_{train}(c_j)} 
\label{eqn:bayes}
\end{equation}

\noindent a unique (up to scale) likelihood probability $\hat{p}(x | c_i)$ solution for each class $c_i$ exists when the original priors  $p_{train}(c_i)$ are known. By cross-multiplying and moving all terms to one side in Eqn.~\ref{eqn:bayes} for the $N$ posterior equations   
\begin{eqnarray}
p(x | c_1)  p_{train}(c_1) - p(c_1 | x)\sum_{j=1}^N\nolimits p(x | c_j) p_{train}(c_j)  & = & 0  \nonumber \\
\vdots \hspace{2.0in}  \\ 
p(x | c_N)  p_{train}(c_N) - p(c_N | x)\sum_{j=1}^N\nolimits p(x | c_j) p_{train}(c_j) & = & 0  \nonumber 
\label{eqn:bayes-linear}
\end{eqnarray}

\noindent a homogeneous linear system of equations can be formulated
\begin{equation}
{\bf M} 
\begin{bmatrix}
p(x | c_1)\\
\vdots\\
p(x | c_N)
\end{bmatrix}
=  \begin{bmatrix}
0\\
\vdots\\
0
\end{bmatrix}
\label{eqn:bayes-matrix}
\end{equation}

\noindent The Perron-Frobenious Theory \cite{Meyer2000} on matrix ${\bf A} = {\bf M} + {\bf I}$ can be used to prove that the unique (up to scale) solution of the likelihoods $\hat{p}(x | c_i)$ is the Eigenvector corresponding to the maximum Eigenvalue $\lambda_{max}$=1 of  the $N$$\times$$N$ matrix ${\bf A}$ (derived in \cite{Davis2020}). Though the likelihoods $\hat{p}(x | c_i )$ are recovered up to scale, their application in Eqn.~\ref{eqn:bayes}  is sufficient, as the ratio cancels any constant scale factor across the likelihoods. 

Given the classifier posteriors $p(c_i | x)$ for a test example $x$ and the original priors $p_{train}(c_i)$, the recovered  likelihoods $\hat{p}(x | c_i )$ and new priors $p_{new}(c_i)$ (provided by the monitoring procedure) can be employed in Eqn.~\ref{eqn:bayes} to yield an updated posterior appropriate for the current Prior Shift context. Thus, though the expected presence of certain classes can change, a trained model can be quickly and efficiently adapted to reflect the new class priors without retraining the original model.


Overall, the ability to be adaptable to changes in the deployment landscape is paramount for continued efficacy of the model. Multiple methods to continually re-align an existing model can be employed to ensure ongoing success. In the specific case of Prior Shift, we presented a robust and provable technique to re-estimate posteriors using the priors of the current situation. Without such adaptability over time, a model may not be flexible enough to new situations and have significantly degraded  performance, thus affecting confidence in the model.


\section{Summary}

In this paper, we proposed 4 tenets of Confident AI: Repeatability, Believability, Sufficiency, and Adaptability. Repeatability employs a statistical summarization and comparison of model performance. Believability uses calibration to best align model confidence to the certainty of prediction. Sufficiency ensures predictions meet a desired level of confidence for critical applications. Adaptability increases the agility of the system to changing situations/environments. Each of the tenets addresses a fundamental issue associated with attributing confidence to AI predictions. Together, they form a principled means for designing and demonstrating system capability with internal (algorithm) and external (user) confidence. The policies presented can be naturally integrated with other complementary AI themes, such as Trustworthy AI, and are also important for AI assurances and verification/auditing. We expect Confident AI to naturally evolve and expand over time, and we encourage AI practitioners to utilize these concepts in developing AI systems.


\section{Ackowledgements}

The supporting work on Confident AI presented in this paper was conducted in collaboration with members of the Ohio State University Computer Vision Laboratory and US Air Force Research Laboratory. 


%

 \bibliographystyle{splncs04}
 \bibliography{mybib}

\begin{thebibliography}{10}
\providecommand{\url}[1]{\texttt{#1}}
\providecommand{\urlprefix}{URL }
\providecommand{\doi}[1]{https://doi.org/#1}

\bibitem{AAAI}
{American Association of Artificial Intelligence}. \url{AAAI.org}, accessed:
  2022-01-05

\bibitem{Aldahdooh2022}
Aldahdooh, A., Hamidouche, W., Fezza, S., Deforges, O.: {Adversarial Example
  Detection for DNN Models: A Review and Experimental Comparison}. Artificial
  Intelligence Review  (2022)

\bibitem{Bishop2006}
Bishop, C.: {Pattern Recognition and Machine Learning}. Springer, New York
  (2006)

\bibitem{Davis2020}
Davis, J.: {Posterior Adaptation With New Priors}. arXiv:2007.01386  (2020)

\bibitem{DavisFrank2021}
Davis, J., Frank, L.: {Revisiting Batch Normalization}. arXiv:2110.13989
  (2021)

\bibitem{Davis2019}
Davis, J., Liang, T., Enouen, J., Ilin, R.: {Hierarchical Semantic Labeling
  With Adaptive Confidence}. In: ISVC (2019)

\bibitem{Davis2021}
Davis, J., Liang, T., Enouen, J., Ilin, R.: {Hierarchical Classification with
  Confidence using Generalized Logits}. In: ICPR (2021)

\bibitem{Deng2012}
Deng, J., Krause, J., Berg, A., Fei-Fei, L.: {Hedging Your Bets: Optimizing
  Accuracy-Specificity Trade-offs in Large Scale Visual Recognition}. In: CVPR
  (2012)

\bibitem{Devore2011}
Devore, J.L.: {Probability \& Statistics for Engineering and the Sciences}.
  Brooks / Cole, Cengage Learning, 8 edn. (2011)

\bibitem{Franc2021}
Franc, V., Prusa, D., Voracek, V.: {Optimal Strategies for Reject Option
  Classifiers}. arXiv:2101.12523  (2021)

\bibitem{Frank2021}
Frank, L., Wiegman, C., Davis, J., Shearer, S.: {Confidence-Driven Hierarchical
  Classification of Cultivated Plant Stresses}. In: WACV (2021)

\bibitem{Geng2021}
Geng, C., Huang, S., Chen, S.: {Recent Advances in Open Set Recognition: A
  Survey}. IEEE Transactions on Pattern Analysis and Machine Intelligence
  \textbf{43}(10),  3614--3631 (2021)

\bibitem{Guo2017}
Guo, C., Pleiss, G., Sun, Y., Weinberger, K.: {On Calibration of Neural
  Networks}. In: ICML (2017)

\bibitem{Hinton2015}
Hinton, G., Vinyals, O., Dean, J.: {Distilling the Knowledge in a Neural
  Network}. In: NIPS Deep Learning and Representation Learning Workshop (2015)

\bibitem{Lee2018}
Lee, K., Lee, K., Min, K., Zhang, Y., Shin, J., Lee, H.: {Hierarchical Novelty
  Detection for Visual Object Recognition}. In: CVPR (2018)

\bibitem{Liang2018}
Liang, X., Zhou, H., Xing, E.: {Dynamic-structured Semantic Propagation
  Network}. In: CVPR (2018)

\bibitem{Mahdavi2021}
Mahdavi, A., Carvalho, M.: {A Survey on Open Set Recognition}. arXiv:2109.00893
   (2021)

\bibitem{Meyer2000}
Meyer, C.: Matrix Analysis and Applied Linear Algebra. SIAM, Philadelphia, PA
  (2000)

\bibitem{Parisi2019}
Parisi, G., Kemker, R., Part, J., Kanan, C., Wermter, S.: {Continual Lifelong
  Learning With Neural Networks: A Review}. Neural Networks  \textbf{113},
  54--71 (2019)

\bibitem{Picard2021}
Picard, D.: {\texttt{torch.manual\_seed(3407)} Is All You Need: On the
  Influence of Random Seeds in Deep Learning Architectures for Computer
  Vision}. arXiv:2109.08203  (2021)

\bibitem{RussellNorvig}
Russell, S., Norvig, P.: Artificial Intelligence: A Modern Approach. Prentice
  Hall, Englewood Cliffs, N.J, 4 edn. (2021)

\bibitem{Tucker2021}
Tucker, P.: {This Air Force Targeting AI Thought it had a 90\% Success Rate. It
  was More Like 25\%}. Defense One  (2021, December 9)

\bibitem{Wu2020}
Wu, T., Morgado, P., Wang, P., Ho, C., Vasconcelos, N.: {Solving Long-tailed
  Recognition with Deep Realistic Taxonomic Classifier}. In: ECCV (2020)

\bibitem{Zhao2020}
Zhao, Y., Chen, J., Oymak, S.: {On the Role of Dataset Quality and
  Heterogeneity in Model Confidence}. In: ICML Uncertainty and Robustness in
  Deep Learning Workshop (2020)

\end{thebibliography}

\end{document}